\documentclass[fleqn,10pt]{wlscirep}
\usepackage[utf8]{inputenc}
\usepackage[T1]{fontenc}
\usepackage{hyperref}
\title{Training-free Ultra Small Model for Universal Sparse Reconstruction in Compressed Sensing}

\author[1,2,*]{Chaoqing Tang}
\author[1]{Huanze Zhuang}
\author[3]{Guiyun Tian}
\author[1]{Zhenli Zeng}
\author[4]{Yi Ding}
\author[1,2]{Wenzhong Liu}
\author[5]{Lin Lin}
\author[6,*]{Xiang Bai}
\affil[1]{School of Artificial Intelligence and Automation, Huazhong University of Science and Technology, Wuhan, China.}
\affil[2]{China Belt and Road Joint Lab on Measurement and Control Technology, Wuhan, China.}
\affil[3]{School of Electric and Electrical Engineering,
Chongqing University of Technology, Chongqing, China.}
\affil[4]{Optics Valley Laboratory, Wuhan, China.}
\affil[5]{School of Water Conservancy and Transportation, Zhengzhou University, Zhengzhou, China.}
\affil[6]{School of Software Engineering, Huazhong University of Science and Technology, Wuhan, China.}

\affil[*]{billtang@hust.edu.cn; xbai@hust.edu.cn}

\begin{abstract}
Pre-trained large models attract widespread attention in recent years, but they face challenges in applications that require high interpretability or have limited resources, such as physical sensing, medical imaging, and bioinformatics. Compressed Sensing (CS) is a well-proved theory that drives many recent breakthroughs in these applications. However, as a typical under-determined linear system, CS suffers from excessively long sparse reconstruction times when using traditional iterative methods, particularly with large-scale data. Current AI methods like deep unfolding fail to substitute them because pre-trained models exhibit poor generality beyond their training conditions and dataset distributions, or lack interpretability. Instead of following the big model fervor, this paper proposes ultra-small artificial neural models called coefficients learning (CL), enabling training-free and rapid sparse reconstruction while perfectly inheriting the generality and interpretability of traditional iterative methods, bringing new feature of incorporating prior knowledges. In CL, a signal of length $n$ only needs a minimal of $n$ trainable parameters. A case study model called CLOMP is implemented for evaluation. Experiments are conducted on both synthetic and real one-dimensional and two-dimensional signals, demonstrating significant improvements in efficiency and accuracy. Compared to representative iterative methods, CLOMP improves efficiency by 100 to 1000 folds for large-scale data. Test results on eight diverse image datasets indicate that CLOMP improves structural similarity index by 292\%, 98\%, 45\% for sampling rates of 0.1, 0.3, 0.5, respectively. We believe this method can truly usher CS reconstruction into the AI era, benefiting countless under-determined linear systems that rely on sparse solution. The code and core data of this paper are available at \url{https://github.com/BillTtzqgbt/CSCoefficientsLearning.git}.
\end{abstract}
\begin{document}

\flushbottom
\maketitle

\thispagestyle{empty}

\section*{Introduction}
Large models show many interesting abilities in recent years, which brings in fervor in both academic and industrial fields. However, large models still lack interpretability and it is resource-intensive \cite{RN1302, 2024SoraAR}, these drawbacks make it difficult to be applied to fields like physical sensing. Compressed Sensing (CS) is a well-proved theory for resource-limited applications, and it is a typical under-determined linear system that integrates sparse representation and sparse reconstruction. Since formally proposed in 2006\cite{RN1283, RN3}, CS has broken the nearly century-long reign of the Nyquist sampling theorem with rigid mathematical proof, leading to various breakthroughs in fields such as engineering imaging with the idea of single pixel imaging \cite{RN930, RN1286, RN374, RN1287, RN1288}, and biomedical imaging like stochastic optical reconstruction microscopy \cite{RN364}, fluorescence imaging \cite{RN1266} and magnetic resonance imaging \cite{RN1264}, genetic engineering \cite{RN1265}, optical imaging \cite{RN563, RN1282}, quantum research \cite{RN1267}, etc. CS significantly reduces sensing time, data volume, and brings other advantage like robust to partial data loss. However, CS suffers from a time- and computationally expensive sparse reconstruction process, which limits its application in many real-world scenarios. Some research efforts have focused on developing specialized hardware to accelerate reconstruction, such as in-memory analog solution \cite{RN1281}. The current three categories \cite{RN1299} of CS/sparse reconstruction methods fail to fully consider the generality, interpretability, efficiency, and accuracy. This is also a  challenge for countless under-determined systems that rely on sparse solution.

The first category is the traditional iterative reconstruction methods. Two representative sub-categories\cite{RN784, RN657} are greedy algorithms, such as Orthogonal Matching Pursuit (OMP) and Iterative Hard Thresholding (IHT), convex optimization techniques, including the Iterative Soft Thresholding Algorithm (ISTA) and the Alternating Direction Method of Multipliers (ADMM). These methods are well-supported by mathematical proof \cite{RN1289,RN1290}, i.e., perfect interpretability. They also exhibit strong generality, as they only require sparsity as a priori knowledge, rather than relying on a training dataset or a fixed measurement matrix. However, a significant drawback is their low efficiency, particularly when dealing with large-scale data, such as long sequences or high-resolution images. Since CS theory is primarily designed for one-dimensional signal, for example, when transfer a 256$\times$256 image into 1D sequence using OMP can take several hours on a consumer-grade CPU.
    
The second category is end-to-end learning \cite{RN1291, RN1296, RN1268, RN1269, RN1297}, which involves constructing a neural network model that directly maps the CS measurement $\bf{y}$ to the target signal $\bf{x}$. With the thriving of deep learning\cite{RN775}, the neural network model can take various forms, including fully connected networks \cite{RN1273}, convolutional neural networks\cite{RN1274, RN1268, RN1278, RN1266}, recurrent neural networks\cite{RN1293, RN1271}, generative adversarial networks\cite{RN1292, RN1272}, Transformer\cite{RN1295}, etc. These methods rely on a large training dataset for effective end-to-end training, so the generality is poor on a different dataset or application. The reconstruction efficiency and accuracy are good after training, but the overall efficiency remains a concern, as the training process is time-consuming and can be less efficient than traditional iterative reconstruction methods. Additionally, these deep learning models are totally black box, which has poor interpretability and reliability \cite{RN1294}.

The third category is deep unfolding that thrives from 2016 \cite{RN936, RN1035} and remains a hot topic in recent research \cite{RN1279, RN1277, RN1275, RN1298}, which enhances the interpretability. These methods construct neural network layers to simulate the iterative rounds of traditional iterative reconstruction method, i.e., unfolding the iterations into sequential neural network layers. Notable examples of these unfolded traditional iterative methods include ISTA-Net \cite{RN938}, ADMM-CSNet \cite{RN1037}, and AMP-Net \cite{RN1038}, and many of the refined versions \cite{RN1029, RN1034}. However, these methods also need a large training dataset, and most of them are specifically designed for images. The training process is time-consuming. When applied to new applications, the generality is significantly inferior to that of traditional iterative-based methods, which limits their practical utility.

In summary, current learning-based methods for CS reconstruction require a model training process and a complex neural model that has a large quantity of trainable parameters. The model must be re-trained whenever there is a change in the measurement rate, measurement matrix, signal dimension, or application. Considering the need for model re-training, these methods are far from efficient. In many applications, it is challenging to collect a dataset that allows these training methods to achieve reasonable reconstruction quality. To advance CS reconstruction to a practical application level in the AI era, inspired by the idea of physics informed neural network \cite{RN1285, RN1250, RN1245}, this paper proposes a training-free artificial neural method called Coefficients Learning (CL) that fully integrates the advantages of traditional iterative reconstruction and neural networks, offering the same generality and interpretability as traditional iterative methods while being significantly faster on large-scale data and more accurate. The implementation of CL on OMP (CLOMP) is used for evaluation on both one-dimensional and tow-dimensional signals.

\section*{Results}
\subsection*{One-dimensional Signal}
\subsubsection*{Evaluation with synthetic 1D data}
\begin{figure}[tb]
\centering
\includegraphics[width=\linewidth]{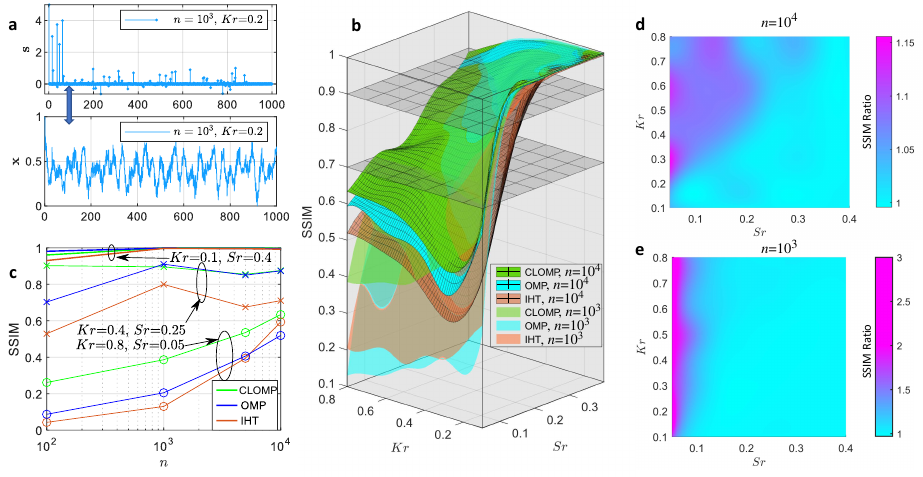}
\caption{\textbf{Reconstruction accuracy based on SSIM in relation to the sparse rate ($Kr$), sampling rate ($Sr$) and signal length ($n$) for 1D signals}. {($\bf{a}$) is an example of a synthetic 1D signal with $n=10^3$ and $Kr=0.2$, where the coefficients $\bf{s}$ and the time-domain signal $\bf{x}$ are presented. The SSIM values for a demo method of the proposed framework (CLOMP) and its mother method OMP, and IHT are presented in ($\bf{b-c}$). ($\bf{c}$) shows the typical best, middle, and worst cases with the combination of $Kr$ and $Sr$ from top to bottom. ($\bf{d,e}$) shows the SSIM ratio of CLOMP to OMP for $n=10^4$ and $n=10^3$, respectively.}}
\label{fig1}
\end{figure}

\begin{figure}[tb]
\centering
\includegraphics[width=\linewidth]{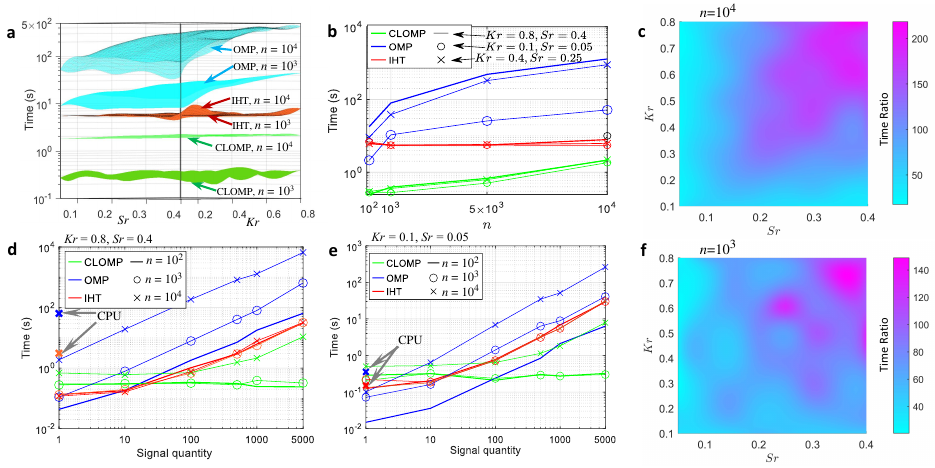}
\caption{\textbf{The reconstruction time of the proposed CLOMP and other methods versus $Sr$, $Kr$, $n$ and signal quantity for 1D signals.} {($\bf{a})$ is a 3D plot for time versus $Sr$ and $Kr$ under different $n$. Time versus $n$ for typical best, middle, and worst cases are denoted with `o', `$\times$', and line in ($\bf{b}$), respectively. ($\bf{d,e}$) show the time versus signal quantity under worst and best cases for the combination of $Sr$ and $Kr$, respectively. The implementation of OMP and IHT under $n=10^4$ on an Intel Xeon Platinum 8352Y CPU is also highlighted for the case where the signal quantity is 1. ($\bf{c,f}$) present the time ratio of CLOMP to OMP for $n=10^4$ and $n=10^3$, respectively.}}
\label{fig2}
\end{figure}

The fundamental model of compressed sensing is $\bf{y} = \bf{Mx}+\bf{\xi}= \bf{MDs}+\bf{\xi} = \bf{As} + \bf{\xi}$, where ${\mathbf{y}} \in {\mathbb{R}^{m \times 1}}$, $\bf{M}$ and ${\mathbf{A}} \in {\mathbb{R}^{m \times n}}$ are the measurements, sensing matrix, and measurement matrix, respectively. $n \gg m$ and $m = \left\lfloor {n{Sr}} \right\rfloor$, where $Sr$ is the sampling rate. ${\mathbf{s}} \in {\mathbb{R}^{n \times 1}}$ is the sparse coefficients, which has $K$ non-zeros values, and the sparse rate is defined as $Kr=K/n$. $\bf{s}$ links to the original signal $\bf{x}$ and sparse basis ${\mathbf{D}} \in {\mathbb{R}^{n \times n}}$ as ${\mathbf{x}} = {\mathbf{Ds}}$, $\bf{\xi}$ is the noise term. So the key factors influencing the reconstruction accuracy of CS are $n$, $Sr$ and $Kr$. To validate the proposed method, this section generates $\bf{x}$ and tests it under various combinations of ($n$, $Sr$,$Kr$) within their typical ranges, with 1000 random signals generated for each combination. Since existing AI methods lack comparable generality, we compare our approach with traditional iterative methods. Among these, greedy algorithms effectively balance accuracy and efficiency. This section takes OMP, a typical greedy algorithm, as an example and modifies it into the Coefficient Learning model, which we refer to as CLOMP. Analyses are then conducted comparing OMP, CLOMP, and another typical greedy algorithm known for its speed with a known $Kr$, IHT. Unless stated otherwise, the normalized cut-off threshold ($T_h$) in OMP and CLOMP for selecting support in each iteration and maximum iteration number ($R$) for all methods, are set to 0.7 and 200, respectively. The results are presented in Figure \ref{fig1} and Figure \ref{fig2}.

To generate $\bf{x}$ under the combination of ($n$, $Kr$), we initialized a length $n$ signal with zeros and randomly selected $K$ positions as the non-zero coefficients. Discrete Cosine Transform (DCT) matrix is chosen as $\bf{D}$ (same for the rest of this paper), so $\bf{s}$ is the frequency components of the signal. A real-world signal typically exhibits dominant values in the low-frequency range and much smaller values in the high-frequency range. To ensure the signals are representative, the sorted non-zero coefficients follow the shape of a Gaussian probability density function, where the low-frequency part contains a small portion of dominant values, accompanied by a long tail distributed in the rest of $\bf{s}$. $\xi$ is considered as small random values in the high-frequency part. An example for ($n=10^3$, $Kr=0.2$) is illustrated in Fig.\ref{fig1}a. $\bf{M}$ is set as random Gaussian matrix (the same notation will be used throughout this paper). To evaluate the similarity between the reconstruction and the ground truth, we utilize several metrics: the Structural Similarity Index Measure (SSIM), Peak Signal-to-Noise Ratio (PSNR), Mean Squared Error (MSE), and Pearson Correlation Coefficient (PCC). SSIM is selected as the primary index because it effectively combines both local and global information. These metrics are calculated for each signal individually before obtaining the mean value for the 1000 signals under each combination. The algorithms are implemented using PaddlePaddle on a Nvidia RTX4090 GPU.

Figure \ref{fig1}(b-e) demonstrates that the proposed method exhibits significantly improved reconstruction quality in low $Sr$ and high $Kr$ regions, achieving approximately 2 to 3 times of SSIM compared to OMP, while maintaining similar accuracy to OMP in other regions. Additional results are presented in Figure \ref{figA1}. Lower values of $Kr$ generally yield better performance across most sampling rates for all methods, as the key of CS recovery lies in sparse reconstruction. Higher values of $Sr$ consistently lead to enhanced performance, as they allow for the extraction of more information from the original signal. A high $Kr$ may result from inadequate sparse representation, signal complexity, or measurement noise, which requires higher $Sr$ to achieve satisfactory performance. The proposed method performs better in these challenging scenarios. Figure \ref{fig1}c demonstrates that a longer $n$ has better performance in most cases, because $\bf{A}$ has more detailed basis and a lower average residual threshold per signal point. The reason why the proposed method has better accuracy is that estimation error in $\bf{s}$ can be corrected with the loss backpropagation. If OMP gets some wrong support positions in $\bf{s}$, these erroneous positions will persist throughout the remaining iterations, leading to error accumulation and progressively deteriorating position estimation. This advantage becomes more significant in low $Sr$ where wrong position in $\bf{s}$ are more likely to be obtained during the early iterations.

In terms of efficiency, on the synthetic dataset with 1000 signals, the proposed method saves 2$\sim$3 orders of magnitude in time consumption than OMP and 1$\sim$2 orders of magnitude than IHT generally, as illustrated in Figure \ref{fig2}(a,c,f). The time of proposed method is not sensitive to $Sr$, $Kr$, $n$ and signal quantity, so the efficiency improvement particularly pronounced in challenging scenarios where these parameters are high. The traditional OMP and IHT exhibit greater efficiency only when recovering a limited number of signals with $n$<1000, as shown in Figure \ref{fig2}(d,e), the proposed method maintains low time consumption in this context while achieving superior accuracy. With the increase of $n$, the proposed method surpasses OMP in efficiency gradually, which demonstrates the efficiency of our method, especially on large-scale data. Example applications for single or several signal with large $n$ are real-time high-resolution sensing\cite{sun2023i}, point-to-point wireless data transfer, recovering high-dimensional signals by reshaping them into 1D shape, etc. Example applications for large signal quantity are wireless sensor networks\cite{ali2022data}, image and other high-dimensional data reconstruction, multi-pass data transfer, and parallel reconstructions.

\subsubsection*{Test on real 1D signals}
\begin{figure}[!h]
\centering
\includegraphics[width=\linewidth]{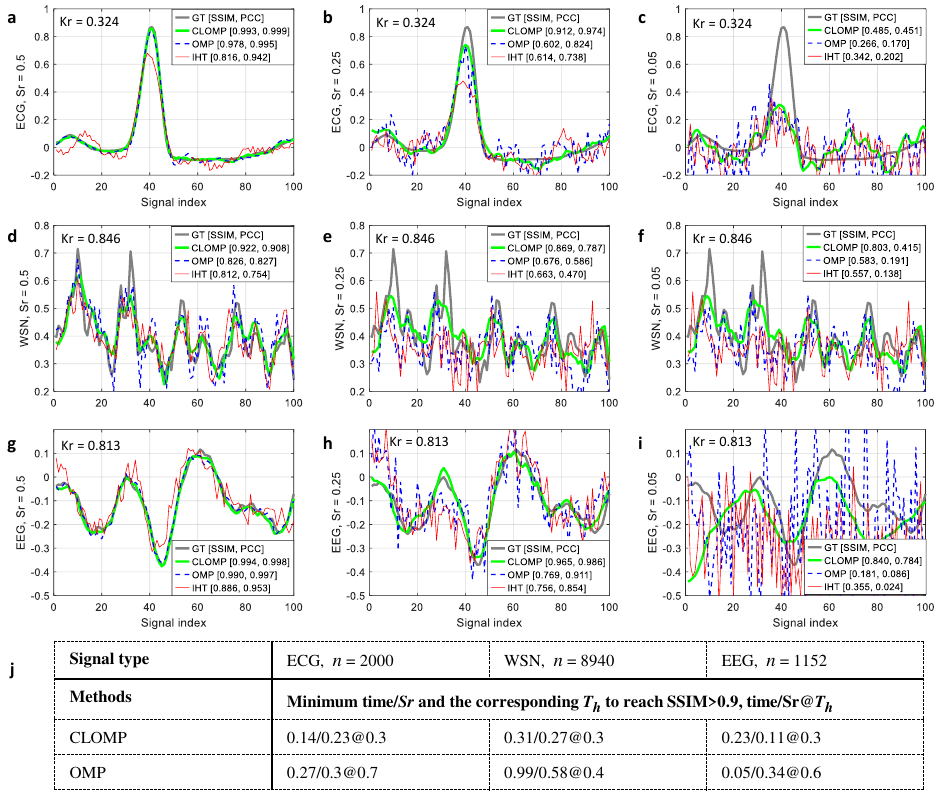}
\caption{\textbf{Examples of reconstructions with a length of 100 on real 1D datasets.}{The ground truths are denoted as GT in ($\bf{a-i}$). The sparse rates ($Kr$) of the full signal length ($n$) are given in the figure. SSIM and PCC of the demo signal are calculated and presented in the square brackets. ($\bf{j}$) provides further evaluation for the proposed CLOMP and its mother method OMP, where the minimum time and $Sr$ to reach SSIM>0.9 are tested for the demo signal in ($\bf{a-i}$), where $T_h$ and maximum iteration number are freely set for each method.}}
\label{fig5}
\end{figure}

One-dimensional signal datasets with varying average sparse rates are examined in this study. The discrete cosine transform (DCT) basis is utilized as $\bf{D}$ again in this section. The $Kr$ is calculated by descending sort the absolute values of the frequency components firstly, and then accumulating the sorted result, if a accumulated value reaches 0.98 of the total summation, $Kr$ is the ratio of the position index to $n$. Average sparse rates are obtained with the average $Kr$ of each signal within a dataset. We selected three datasets from electrocardiogram (ECG), electroencephalography (EEG), and wireless sensor networks (WSN), because compressed sensing has significant potential in these applications for conserving power or compress data of sensor greatly. More specifically, implantable biomedical devices like cardiac pacemaker are limited in power and difficult to charge. These devices usually equip with low-power sensors to capture signals like ECG and EEG. With CS, these implantable sensors only need to wake up for a brief period during each cycle, resulting in substantial power savings. Similar advantages are also appealing for WSN. So, this paper selects Physionet2017ECG\footnote{https://www.kaggle.com/datasets/luigisaetta/physionet2017ecg}, Slow Cortical Potentials\footnote{https://www.kaggle.com/datasets/towsifahamed/eeg-dataset-of-slow-cortical-potentials}, and AirQuality\footnote{https://www.kaggle.com/datasets/fedesoriano/air-quality-data-set} for ECG, EEG, and WSN, their signal lengths are shown in Figure \ref{fig5}j, their signal quantity are 8529, 2660, and 12, respectively. To save time, if the signal quantity is more than 1000, this study only takes the first 1000 signals. The test results are shown in Figure \ref{fig5} and Table \ref{tab1}.

A piece of reconstructed signal with a length of 100 from the three datasets is shown in Figure \ref{fig5}(a-i). The ECG signal is smooth, exhibiting a good $Kr$ of 0.324, the WSN signal is more fluctuating, with a $Kr$ of 0.846, making it more challenging to reconstruct. The EEG signal falls between the two. These sub-figures show SSIM is  more strict to define perfect reconstruction than PCC and consider both local and global similarity well. This is why SSIM is chosen as the primary metric. For instance, the OMP reconstruction near the signal index of 50$\sim$60 has obvious local difference from the ground truth, PCC gets a value of 0.995 while the proposed CLOMP gets a lower value of 0.978. At a high sampling rate of 0.5, only the proposed method achieves an SSIM greater than 0.9 of SSIM on all signals, while IHT exhibits the worst performance. Under a low $Sr$ of 0.05, the performance improvement of the proposed method is particularly pronounced. Figure \ref{fig5}i shows OMP and IHT almost failed, but the proposed method still gets the overall shape of the ground truth. Due to better accuracy, the proposed method requires a lower $Sr$ to reach SSIM>0.9. In resource-limited applications like WSN, lower $Sr$ means more saving in data and power consumption. In medical sensing applications, a lower $Sr$ results in reduced exposure to harmful X-rays for patients. For the three demo signals, the proposed method reduces $Sr$ by 0.07, 0.31, 0.23, corresponding to $-23.3\%$, $-53.4\%$, $-67.6\%$ of reduction in measurement rates.

\begin{table}[htb]
  \centering
  \caption{{\bf{Test results on real 1D datasets}}. {The mean sparse rate is denoted as $\bar Kr$. The improvement are calculated by comparing the proposed CLOMP and best one among OMP and IHT.}}
  
  \resizebox{1\textwidth}{!}{
  \begin{tabular}{rlp{15.835em}p{8.29em}p{8.75em}p{8.625em}p{8.96em}}
    \toprule
    \multicolumn{1}{l}{\textbf{Dataset}} & \textbf{Sr} & \multicolumn{1}{l}{\textbf{Time (s)}} & \multicolumn{1}{l}{\textbf{Mean SSIM}} & \multicolumn{1}{l}{\textbf{Mean MSE}} & \multicolumn{1}{l}{\textbf{Mean PCC}} & \multicolumn{1}{l}{\textbf{Mean PSNR}} \\
    \midrule
     \multicolumn{1}{l}{Type} &   & CLOMP, OMP-CPU/GPU& \multicolumn{4}{p{34.625em}}{\centering \hspace{2.3cm} CLOMP/OMP/IHT} \\
$\bar Kr$ &  &  IHT-CPU/GPU & \multicolumn{4}{p{34.625em}}{\centering \hspace{2.3cm} Improvement (\%)} \\
    \midrule
    \multicolumn{1}{l}{ECG} & 0.05 & 0.227, 6.986/26.661\newline{}20.753/6.716 & 0.549/0.344/0.417\newline{}31.65\% & 0.026/0.055/0.039\newline{}-33.3\% & 0.519/0.222/0.231\newline{}124.68\% & 16.239/13.556/14.435\newline{}12.50\% \\
\cmidrule{2-7}   0.549    & 0.25 & 0.314, 164.246/136.008\newline{}25.791/6.256 & 0.903/0.628/0.657\newline{}37.44\% & 0.003/0.015/0.015\newline{}-80.0\% & 0.955/0.774/0.738\newline{}23.39\% & 26.379/19.201/18.846\newline{}37.38\% \\
\cmidrule{2-7}       & 0.5 & 0.319, 344.126/197.088\newline{}35.563/6.157 & 0.991/0.970/0.847\newline{}2.16\% & 0.0002/0.0006/0.004\newline{}-66.67\% & 0.997/0.989/0.927\newline{}0.81\% & 37.622/33.359/24.311\newline{}12.78\% \\
    \midrule
    \multicolumn{1}{l}{WSN} & 0.05 & 0.319, 0.814/0.591\newline{}0.990/0.191 & 0.704/0.476/0.427\newline{}47.90\% & 0.023/0.022/0.031\newline{}4.55\% & 0.363/0.356/0.284\newline{}1.97\% & 16.918/16.807/15.404\newline{}0.66\% \\
\cmidrule{2-7}  0.897     & 0.25 & 0.312, 23.214/6.188\newline{}3.554/0.170 & 0.812/0.613/0.604\newline{}32.46\% & 0.006/0.012/0.015\newline{}-50.0\% & 0.811/0.664/0.583\newline{}22.14\% & 22.669/19.522/18.545\newline{}16.12\% \\
\cmidrule{2-7}       & 0.5 & 0.327, 100.919/19.082\newline{}13.136/0.149 & 0.886/0.790/0.759\newline{}16.73\% & 0.003/0.005/0.007\newline{}-40.0\% & 0.917/0.860/0.797\newline{}6.63\% & 26.611/23.898/22.278\newline{}11.35\% \\
    \midrule
    \multicolumn{1}{l}{EEG} & 0.05 & 0.476, 3.705/28.882\newline{}17.057/7.956 & 0.735/0.254/0.404\newline{}81.93\% & 0.029/0.099/0.066\newline{}-56.06\% & 0.614/0.275/0.280\newline{}119.29\% & 15.886/10.985/12.250\newline{}29.68\% \\
\cmidrule{2-7}  0.751     & 0.25 & 0.276, 66.716/93.758\newline{}24.791/8.056 & 0.921/0.712/0.645\newline{}29.35\% & 0.004/0.013/0.017\newline{}-69.23\% & 0.955/0.839/0.722\newline{}13.83\% & 24.799/20.080/18.779\newline{}23.50\% \\
\cmidrule{2-7}       & 0.5 & 0.269, 133.686/137.207\newline{}25.864/8.073 & 0.980/0.977/0.865\newline{}0.31\% & 0.001/0.001/0.005\newline{}-10.34\% & 0.992/0.991/0.923\newline{}0.10\% & 31.590/31.193/24.250\newline{}1.27\% \\
    \bottomrule
    \end{tabular}}
  \label{tab1}%
\end{table}%

Table \ref{tab1} presents the average performance on these datasets. The proposed CLOMP demonstrates significantly lower time consumption compared to the original OMP on all datasets. IHT is the most efficient on WSN dataset, because the time required for the GPU implementation of IHT is primarily influenced by the quantity of signals, as illustrated in Figure \ref{fig2}, while WSN dataset only 12 signals. In most cases, the time of CPU implementation is much longer than it's GPU implementation, except for OMP in the EEG dataset and the low $Sr$ regions for ECG dataset, because the signal length or average $Kr$ for these two exceptions is relatively small, resulting in the time for tensor generation and data transfer exceeding that of parallel computing. Regarding accuracy, the proposed method consistently outperforms others across all metrics, with significant improvements noted in low $Sr$ cases. For example, the proposed CLOMP achieves approximately 32\%, 48\%, and 82\% of improvement in SSIM under a $Sr$ of 0.05 for the three datasets, respectively.

\subsection*{Two-dimensional Signal}
\subsubsection*{Evaluation with synthetic 2D data}
One popular method for reconstructing images and high-dimensional signals is reshaping the image into 1D signal, but the signal length will be excessively large. For example, one 128$\times$128 image generates a signal length of $n=128\times 128=16384$, the $\bf{D}$ and reconstruction time increase significantly with higher image resolution. Another way is regarding them as a large collection of 1D signals, where each signal has a different $\bf{s}$ and corresponds to different columns in $\bf{A}$, so the pseudo-inverse operation for traditional reconstruction algorithm like OMP need to prepare different input for each row or column, which is slow. Another drawback of traditional iterative methods is their inability to incorporate prior information, such as local smoothness. The proposed method makes $\bf{s}$ become learnable parameters, thereby bypassing the need for the pseudo-inverse operation. This allows for the integration of various types of prior information into the model through loss backpropagation.

\begin{figure}[tb]
\centering
\includegraphics[width=\linewidth]{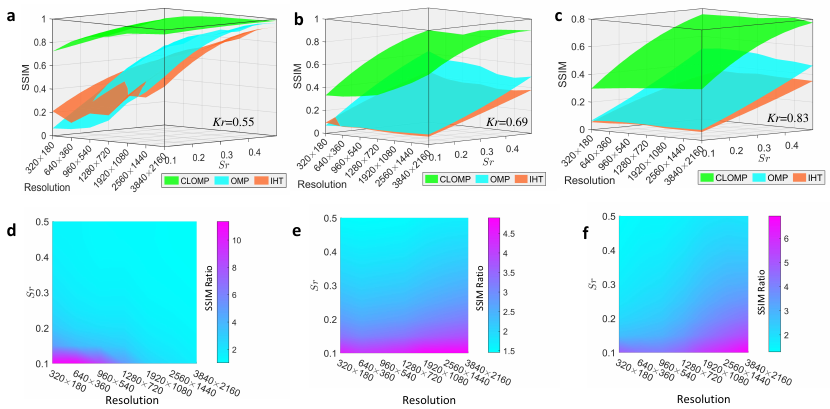}
\caption{\textbf{The reconstruction accuracy based on SSIM in relation to image resolution, $Sr$ and $Kr$ for 2D signals.} {($\bf{a-c}$) are for the synthetic images with $Kr$=0.55, 0.69, 0.83, respectively. ($\bf{d-f}$) are the corresponding SSIM ratios of CLOMP to OMP or $Kr$=0.55, 0.69, 0.83, respectively.}}
\label{fig3}
\end{figure}

\begin{figure}[htb]
\centering
\includegraphics[width=\linewidth]{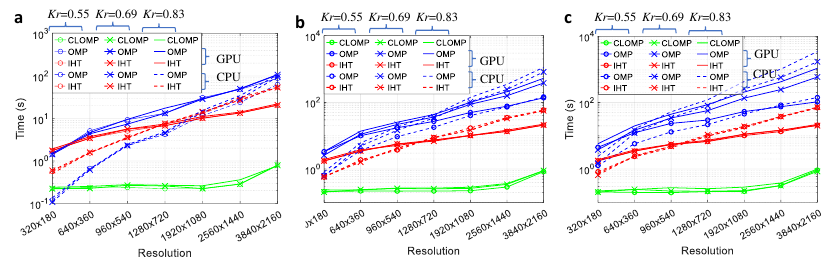}
\caption{\textbf{Reconstruction time versus image resolution, $Sr$ and $Kr$ for 2D signals}. {($\bf{a-c}$) are for low, middle, and high sampling rates with $Sr$=0.1, 0.3, 0.5, respectively.}}
\label{fig4}
\end{figure}

To evaluate the performance on 2D data under important CS settings like $Sr$, $Kr$, $n$ and signal quantity, this section utilizes three 4K resolution images with low, middle and high $Kr$, and resizes them into different resolutions, because $n$ and signal quantity are the row and column numbers of an image, i.e., the resolution. A group of images is presented in Figure \ref{figA2}. We find smaller resolution for the same image usually has higher $Kr$, to make the resized images keep the same $Kr$, Gaussian smoothing with varying parameters is applied. The parameter $Sr$ is set within its typical range of 0.1 to 0.5. Grayscale images are employed for this analysis. In this test, local variation and total variation are utilized as prior information. The results are summarized in Figure \ref{fig3} and Figure \ref{fig4}.

The accuracy of the proposed method is at least twice as high as that of OMP in most cases, with improvements ranging from four to ten times under low $Sr$, as is shown in Figure \ref{fig3}(d-f). Figure \ref{fig3}(a-c) demonstrates that a lower sparsity rate generally leads to decreased accuracy. For instance, with an image resolution of 320$\times$180 and $Sr=0.1$, the three images get SSIM of 0.72, 0.35, 0.30 for their $Kr$ of 0.55, 0.69, 0.83, respectively. With the increase of image resolution or $Sr$, the accuracy becomes better, and the influence of $Sr$ is more significant than image resolution. This figure also indicates that the proposed method requires significantly less $Sr$ to reach the same recovery accuracy, which can greatly reduce the time during data acquisition, indicating great potential in medical imaging.

In terms of efficiency, Figure \ref{fig4} indicates that other methods only possibly more efficient for super low resolution images, but a high-enough image resolution is important for almost all applications. In many scenarios, the proposed method is 2 to 3 orders of magnitude more efficient than traditional iterative algorithms. For example, the proposed method can recover a 4K image in less than 1 second, while OMP needs at least approximately 100 seconds on GPU and around 1000 seconds on CPU for $Sr$>0.3. The time of the proposed method is minimally affected by changes in resolution, whereas the processing time of other methods increases significantly with resolution. Another observation is that images with larger $Kr$ typically require more time to reconstruct, but the influence is not so significant as resolution, so a good sparse representation will enhance both accuracy and efficiency.

\subsubsection*{Test on image datasets with various resolutions, modalities, and sparse rates}
\begin{figure}[!ht]
\centering
\includegraphics[width=1\linewidth]{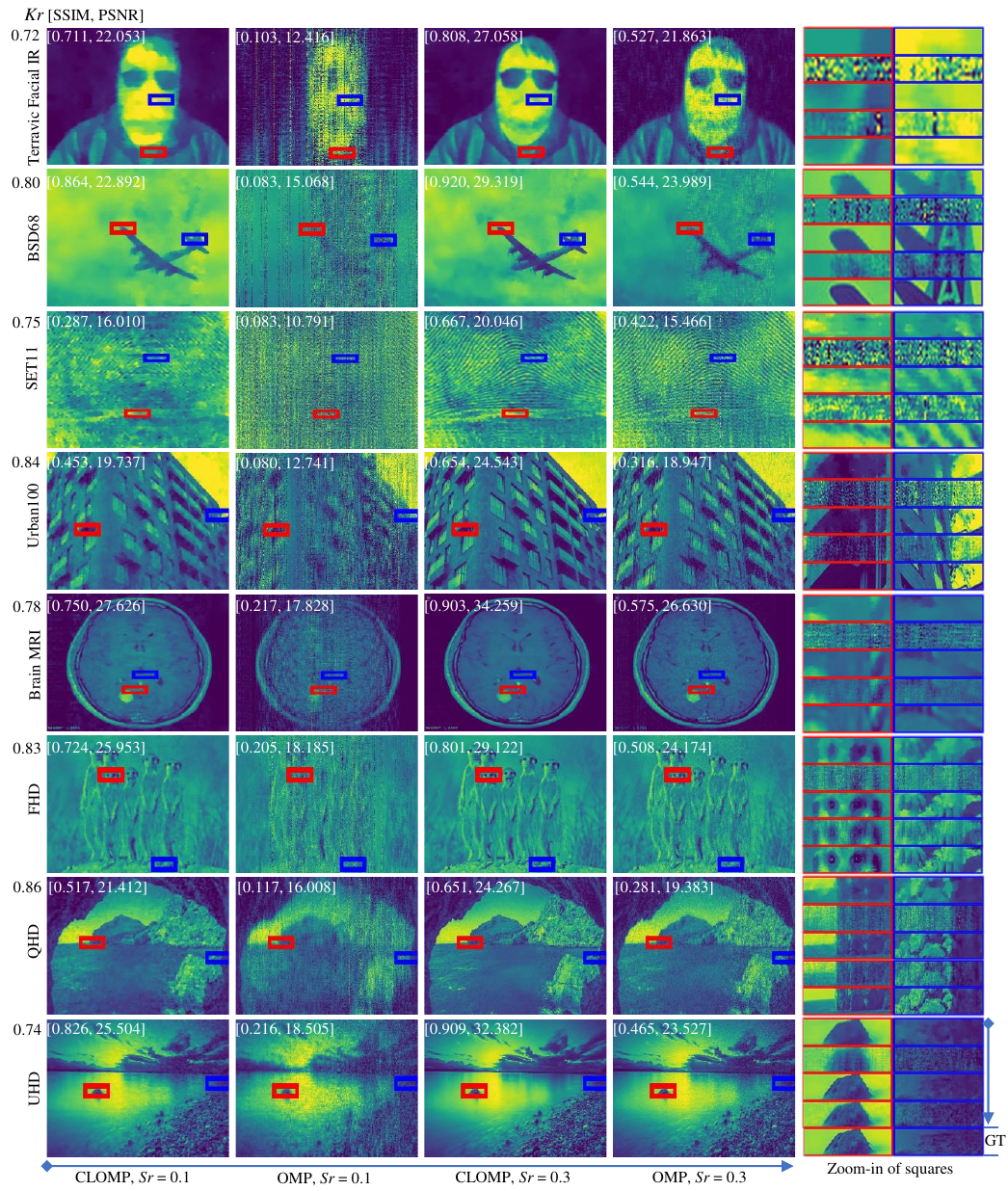}
\caption{\textbf{Reconstruction examples.} {The zoom-in of squares are displayed in the rightmost column, where the bottom row is from the ground truth (GT) images. The image resolutions are shown in Table \ref{tab2}.}}
\label{figImg}
\end{figure}

This section evaluates the proposed method on various image datasets and our collected datasets for CS with varying resolutions, modalities, and sparse rates. To maintain the same number of images as in Set 11, we also selected 11 images from other dataset as well , which is sufficient to demonstrate the accuracy and efficiency of our method. We collected additional images with FHD and UHD resolution, so the image datasets covers the resolution from $320\times 240$ to $3840\times 2160$, the modalities includes infrared, visible light, MRI. The full dataset can be found \href{https://github.com/BillTtzqgbt/CSCoefficientsLearning.git}{here}. We employed well-known metrics like SSIM, MSE, PSNR. The detailed information like resolution and average sparse rates for the datasets is provided in Table \ref{tab2}. One image from each dataset is displayed in Figure \ref{figImg}. Note that we did not exhaustively optimize the proposed method during this test, adjusting the weight parameters may get better accuracy in these results. With a $Sr$ of 0.1, the proposed method has 0.5, 7.5 of improvements on average on SSIM and PSNR for these images, respectively. At a $Sr$ of 0.3, the average improvement still reach 0.33 and 5.87, respectively. Generally, a lower $Kr$ leads to better accuracy. The zoom-in squares on the right column demonstrate that the proposed method effectively recovers both the abstract shape and fine details under $Sr$=0.3 for all datasets. Since local variance is utilized as prior information in this study, images with a smooth local appearance exhibit better accuracy, such as the image selected from BSD68. Conversely, images with fluctuating local regions get lower accuracy with the proposed method, as seen in the fingerprint image selected from SET11. Both total variance and local variance are universal prior information for images that can be applied in most of scenes. In practical applications, users can based on these two priors and incorporate additional prior information into their loss function.

More comprehensive test results on these datasets are presented in Table \ref{tab2}. Compared to OMP and IHT, the proposed method has 292\%, 98\%, 45\% of improvement on average on SSIM for $Sr$=0.1, 0.3, 0.5 respectively, and has 46\%, 28\%, 17\% of improvement on average on PSNR. The MSE improvement are more consistent across all $Sr$ with an average reduction of 67\%. Thus, the proposed method significantly outperforms traditional iterative algorithms in accuracy. In terms of efficiency, the proposed CLOMP only 76 microseconds slower than the CPU implementation of OMP under $Sr$=0.1 for the resolution of 320$\times$240. It is much faster than other iterative algorithms for higher resolution images and higher $Sr$. On the 4K resolution dataset, by substituting the pseudo-inverse with learnable parameters, CLOMP is 117 and 29 times faster than OMP and IHT with $Sr$=0.1, respectively; 552 and 23 times faster with $Sr$=0.5. With the increase of image resolution, the time improvements become even more significant. MRI, infrared images and 4K images perform better than other scenes, because they have relatively lower sparse rate on DCT.

\begin{table}[t]
  \centering
  \caption{{\bf{Test results on diverse image datasets}}. {The improvement are calculated by comparing the proposed CLOMP and best one among OMP and IHT.}}
  
  \resizebox{1\textwidth}{!}{
   
    \begin{tabular}{cllp{7.375em}p{6.29em}p{12.125em}p{12.25em}p{13.29em}}
    \toprule
    \multicolumn{1}{l}{\textbf{Dataset}} & \textbf{Sr} & \multicolumn{3}{l}{\textbf{Time (s)}} & \multicolumn{1}{l}{\textbf{Mean SSIM}} & \multicolumn{1}{l}{\textbf{Mean MSE}} & \multicolumn{1}{l}{\textbf{Mean PSNR}} \\
    \midrule
    \multicolumn{1}{p{5.585em}}{\small Name\newline{}    \text{     } Resolution\newline{}  \text{     }   Mean $Kr$} & \multicolumn{4}{p{18.165em}}{\flushright \small CLOMP|OMP-CPU/GPU|IHT-CPU/GPU} & \multicolumn{3}{p{37.665em}}{\centering \vspace{0.4cm} \small CLOMP/OMP/IHT, Improvement(\%)} \\
    \midrule
    \multicolumn{1}{l}{Facial IR} & 0.1 & \multicolumn{1}{r}{0.256 } & 0.180/1.984 & 0.611/1.841 & 0.707/0.169/0.233, 203.43\% & 0.009/0.065/0.043, -79.07\% & 21.818/12.119/13.953, 56.37\% \\
\cmidrule{2-8}    320$\times$240 & 0.3 & 0.227  & 0.992/4.304 & 0.702/1.850 & 0.844/0.498/0.478, 69.48\% & 0.002/0.005/0.009, -60.00\% & 28.663/23.252/20.797, 23.27\% \\
\cmidrule{2-8}    0.707 & 0.5 & 0.210  & 1.585/5.542 & 0.743/1.843 & 0.877/0.729/0.655, 20.30\% & 0.001/0.001/0.003, -21.43\% & 30.608/28.834/26.006, 6.15\% \\
    \midrule
    \multicolumn{1}{l}{BSD68} & 0.1 & 0.260  & 0.418/2.874 & 1.733/2.590 & 0.507/0.056/0.122, 315.57\% & 0.015/0.075/0.050, -70.00\% & 19.531/11.578/13.669, 42.89\% \\
\cmidrule{2-8}    481$\times$321 & 0.3 & 0.258  & 3.073/7.667 & 1.883/2.609 & 0.665/0.293/0.268, 126.96\% & 0.006/0.020/0.022, -70.00\% & 23.956/18.113/17.592, 32.26\% \\
\cmidrule{2-8}    0.84 & 0.5 & 0.260  & 7.681/11.753 & 2.001/2.608 & 0.761/0.475/0.375, 60.21\% & 0.003/0.009/0.012, -66.67\% & 26.625/21.592/20.434, 23.31\% \\
    \midrule
    \multicolumn{1}{l}{SET11} & 0.1 & 0.264  & 0.920/3.853 & 2.740/2.487 & 0.471/0.045/0.085, 454.12\% & 0.016/0.093/0.056, -71.43\% & 18.525/10.403/12.829, 44.40\% \\
\cmidrule{2-8}    512$\times$512 & 0.3 & 0.267  & 6.237/10.147 & 3.344/2.492 & 0.674/0.275/0.234, 145.09\% & 0.006/0.022/0.024, -72.73\% & 23.028/17.123/16.540, 34.49\% \\
\cmidrule{2-8}    0.821 & 0.5 & 0.265  & 16.884/16.558 & 3.790/2.487 & 0.777/0.490/0.373, 58.57\% & 0.003/0.009/0.014, -66.67\% & 25.627/21.104/19.017, 21.43\% \\
    \midrule
    \multicolumn{1}{l}{Urban100} & 0.1 & 0.266  & 3.831/12.840 & 8.438/5.590 & 0.465/0.054/0.100, 365.00\% & 0.016/0.084/0.061, -73.77\% & 18.744/10.996/12.992, 44.27\% \\
\cmidrule{2-8}    1024$\times$768 & 0.3 & 0.272  & 29.515/34.193 & 11.489/5.558 & 0.651/0.282/0.237, 130.85\% & 0.006/0.022/0.026, -72.73\% & 23.008/17.592/16.807, 30.79\% \\
\cmidrule{2-8}    0.827 & 0.5 & 0.276  & 80.616/61.679 & 12.740/5.619 & 0.763/0.450/0.359, 69.56\% & 0.003/0.011/0.015, -72.73\% & 25.585/20.501/19.034, 24.80\% \\
    \midrule
    \multicolumn{1}{l}{Brain MRI} & 0.1 & 0.222  & 6.145/16.930 & 10.206/5.517 & 0.767/0.232/0.244, 214.34\% & 0.003/0.025/0.029, -88.00\% & 26.000/16.323/15.844, 59.28\% \\
\cmidrule{2-8}    1024$\times$1024 & 0.3 & 0.223  & 35.376/39.706 & 13.135/5.534 & 0.853/0.570/0.462, 49.65\% & 0.001/0.003/0.006, -66.67\% & 33.368/26.258/22.471, 27.08\% \\
\cmidrule{2-8}    0.645 & 0.5 & 0.217  & 49.809/45.137 & 13.366/5.476 & 0.926/0.847/0.689, 9.33\% & 0.001/0.001/0.002, -40.00\% & 36.460/33.965/27.997, 7.35\% \\
    \midrule
    \multicolumn{1}{l}{FHD} & 0.1 & 0.263  & 13.746/32.857 & 21.950/10.548 & 0.536/0.085/0.139, 285.61\% & 0.009/0.042/0.031, -70.97\% & 21.727/14.138/15.770, 37.77\% \\
\cmidrule{2-8}    1920$\times$1080 & 0.3 & 0.279  & 101.367/87.383 & 27.720/10.514 & 0.674/0.348/0.304, 93.68\% & 0.004/0.013/0.015, -69.23\% & 25.292/20.213/19.345, 25.13\% \\
\cmidrule{2-8}    0.844 & 0.5 & 0.268  & 291.367/159.825 & 27.498/10.471 & 0.764/0.499/0.420, 53.11\% & 0.003/0.007/0.009, -57.14\% & 27.329/23.235/21.785, 17.62\% \\
    \midrule
    \multicolumn{1}{l}{QHD} & 0.1 & 0.283  & 21.335/40.507 & 30.523/11.302 & 0.552/0.100/0.137, 302.92\% & 0.007/0.033/0.029, -75.86\% & 22.316/14.902/15.607, 42.99\% \\
\cmidrule{2-8}    2040$\times$1356 & 0.3 & 0.316  & 167.056/117.493 & 32.142/11.289 & 0.705/0.339/0.294, 107.96\% & 0.003/0.009/0.012, -66.67\% & 26.420/20.909/19.770, 26.36\% \\
\cmidrule{2-8}    0.825 & 0.5 & 0.320  & 468.523/210.805 & 32.892/11.262 & 0.792/0.507/0.424, 56.21\% & 0.002/0.005/0.007, -60.00\% & 28.803/24.051/22.297, 19.76\% \\
    \midrule
    \multicolumn{1}{l}{UHD} & 0.1 & 0.718  & 84.276/107.045 & 54.510/20.857 & 0.707/0.219/0.234, 202.14\% & 0.005/0.023/0.024, -78.26\% & 25.813/17.980/17.970, 43.57\% \\
\cmidrule{2-8}    3840$\times$2160 & 0.3 & 0.863  & 631.441/318.513 & 57.581/20.703 & 0.831/0.514/0.467, 61.67\% & 0.002/0.007/0.008, -71.43\% & 31.496/25.057/23.305, 25.70\% \\
\cmidrule{2-8}    0.777 & 0.5 & 0.915  & 1484.288/505.072  & 61.335/20.962 & 0.880/0.683/0.606, 28.84\% & 0.001/0.003/0.004, -66.67\% & 33.464/28.866/26.200, 15.93\% \\
    \bottomrule
    \end{tabular}%
  }
  \label{tab2}%
\end{table}%

\section*{Discussion}
This paper proposes ultra-small artificial neural models called coefficients learning (CL), enabling training-free and rapid sparse reconstruction while perfectly inheriting the generality and interpretability of traditional iterative methods. CL is a new category is CS reconstruction methods, which brings additional advantage of incorporating  prior knowledges. In CL, a signal of length $n$ only needs a minimal of $n$ trainable parameters. For example, reconstructing a signal with length $n=10$, CL only has 10 trainable parameters, which is ultra small. Current AI-based CS reconstruction methods often fail when there are changes in sampling rates, measurement matrices, signal dimensions, or applications. The efficiency are 2$\sim$3 orders of magnitude higher than that of traditional iterative algorithms for large-scale data. The accuracy on eight representative image datasets demonstrates an average improvement of 292\%, 98\%, 45\% on SSIM for $Sr$=0.1, 0.3, 0.5, respectively. The perfect generality, interpretability, and good efficiency, accuracy enable CL to effectively replace traditional iterative algorithms for sparse reconstruction. This is particularly beneficial for applications where sensing time is important, such as  resource-limited sensors in wireless sensor networks, implantable biomedical devices, and medical imaging. Since compressed sensing is typical under-determined linear system, the proposed CL method also works for countless applications that require sparse solution of under-determined linear systems, such as image super-resolution, modeling of electromagnetic scattering, blind signal separation.

We find that many iterative algorithms (e.g. OMP, IHT, SP, ISTA) for sparse reconstruction can be divided into two distinct blocks. The first block is responsible for extracting new information from the residual and injecting the information into the new estimation, while the second block updates the estimation. The first block is to guarantee the 0-norm in CS model, the second block is to update the $\bf{s}$ and residual using a complex and time-consuming equation. So, this paper proposes a framework called coefficients learning, which consists of a residual injection branch and a $\bf{s}$ forward branch to substitute tradition iterative methods. The residual injection branch reformulates the corresponding component in the iterative algorithm by utilizing fixed parameters in neural layers and activation functions. If the branch is not overly complex, reformulating it into GPU implementation is also an efficient choice. For the second block, we directly set $\bf{s}$ as learnable parameters to bypass the time-consuming equation. We propose several branch structures for $\bf{s}$ learning. The advantage of using a learnable $\bf{s}$ is that prior information can be integrated into the loss function to enhance performance. We proposes the total variation and local variation as two universal prior loss terms for both 1D and high-dimensional signals. Another significant advantage for CL is the number of learnable parameters is extremely low compared to other AI method, because only $n$ parameters are required in a minimal structure, resulting in a rapid learning process.

To evaluate the proposed method, this paper takes OMP as a case study and reformulates it into CL structure as CLOMP. Tests are carried out in both synthetic and real application 1D and 2D signals, comparing CLOMP with its parent algorithm OMP, and a fast iterative algorithm IHT. The synthetic data is utilized to assess performance based on key parameters in CS, e.g. $Kr$, $Sr$, $n$. Evaluation on synthetic 1D data shows the proposed CLOMP gets 2$\sim$3 times of SSIM than OMP in low $Sr$ and high $Kr$ region, and has similar accuracy in other regions. In terms of efficiency for 1D signals, OMP and IHT are only more efficient in recovering several signals with $n$<1000. The time of CLMOP is not sensitive to $Sr$, $Kr$, $n$ and signal quantity, so with the improvement of these parameters, it saves 2$\sim$3 orders of magnitude of time than OMP and 1$\sim$2 orders of magnitude than IHT generally under a signal quantity of 1000. Tests on real ECG, EEG, WSN data corroborate these findings, and CLOMP saves $23.3\%$, $53.4\%$, $67.6\%$ of measurement rates to reach SSIM>0.9, indicating great potential in saving power for sensors and minimizing harmful radiation exposure for patients in medical sensing applications. 

Tests on synthetic images demonstrate that the accuracy of CLOMP is at least twice as high as that of OMP in most cases, with improvements ranging from four to ten times under low $Sr$. Other methods only possibly more efficient for super low resolution images. In other situations, the CLOMP is two to three orders of magnitude more efficient than traditional iterative algorithms. Eight image datasets that have resolution from 320$\times$240 to 3840$\times$2160 with different modalities are used for real 2D signal evaluation. The results show 292\%, 98\%, 45\% of improvement on average on SSIM for $Sr$=0.1, 0.3, 0.5 respectively, and 46\%, 28\%, 17\% of improvement on average on PSNR. On the 4K resolution dataset, CLOMP is 117 and 29 times faster than OMP and IHT with $Sr$=0.1, respectively; 552 and 23 times faster with $Sr$=0.5 .

There are also some limitations that worth to be addressed in future work. Firstly, the weight factors for prior information are empirically set currently, a bad setting can decrease accuracy. We find these weight factors are linked to $Kr$, $Sr$ and signal features. A high $Sr$ needs lower priori weight. An image with smooth regions necessitates a higher weight. However, a quantitative guiding rule has not yet been established. Secondly, the TV and LV loss needs to calculate $\bf{D}s$, when $n$ is large, ${\mathbf{D}} \in {\mathbb{R}^{n \times n}}$ will occupy too many memory space, which limits efficiency. This explains why the efficiency decreases, and consequently, the time required for CLOMP increases as $n$ grows.

\section*{Methods}
\subsection*{Mathematic Model}
The basic model of compressed sensing is designed for 1D signal ${\mathbf{x}} \in {\mathbb{R}^{n \times 1}}$ to get a measurement ${\mathbf{y}} \in {\mathbb{R}^{m \times 1}}$, $n \gg m$ so the measurements are compressed comparing to the original signal. So high-dimensional data is usually reshaped into 1D. The model is,
\begin{equation}
    \bf{y} = \bf{Mx}+\bf{\xi}=\bf{MDs}+\bf{\xi}=\bf{As} + \bf{\xi}
    \label{eq1}
\end{equation}
where ${\mathbf{M}} \in {\mathbb{R}^{m \times n}}$ and ${\mathbf{A}} \in {\mathbb{R}^{m \times n}}$ are the sensing matrix and measurement matrix, respectively. ${\mathbf{s}} \in {\mathbb{R}^{n \times 1}}$ is the sparse coefficients, which is the sparse representation of $\bf{x}$ on the sparse basis ${\mathbf{D}} \in {\mathbb{R}^{n \times n}}$, $\bf{\xi}$ is the noise term.  The reconstruction process is solving the following problem,
\begin{equation}
    \min {\left\| {\mathbf{s}} \right\|_0}\quad s.\:t.{\text{\;}}{\left| {\left| {{\mathbf{As}} - {\mathbf{y}}} \right|} \right|_2} < \varepsilon
    \label{eq2}
\end{equation}
where ${\left\| {\mathbf{\cdot}} \right\|_0}$ means the 0-norm. So CS reconstruction is use as less components in $\bf{s}$ as possible the make the residual $\bf{r}={\mathbf{As}} - {\mathbf{y}}$ less than a threshold, $\varepsilon$. Methods like OMP, IHT, ISTA can be used to solve this problem without training. For a group of signals like image, the original signal ${\mathbf{X}} \in {\mathbb{R}^{n \times a}}$ and measurement results ${\mathbf{Y}} \in {\mathbb{R}^{m \times a}}$ are matrices, each column of $\bf{X}$ usually share the same measurement matrix as is shown in Eq.(\ref{eq3}). 
\begin{equation}
    {\mathbf{Y}} = {\mathbf{MX}} + {\mathbf{\xi }} = {\mathbf{MDS}} + {\mathbf{\xi }} = {\mathbf{AS}} + {\mathbf{\xi }}
    \label{eq3}
\end{equation}

This paper introduces a new reconstruction model as presented in Eq.(\ref{eq4}), where ${\mathbf{S}}_i$ represents the $i$-th column in ${\mathbf{S}}$. One-dimensional signal is a special case of this model where $a=1$. In contast to traditional reconstruction problems, prior information is integrated into this model. This information is application-dependent and may include known features of $\bf{x}$ or $\bf{s}$. This paper proposes two universal prior information: total variation (TV) and local variation (LV). Total variation for 1D signal and images are given in Eq.(\ref{eq5}) and Eq.(\ref{eq6}) respectively, where $H$ and $W$ are the pixel number in height and width of the image. TV suppresses outliers and ensures that the reconstructed signal is smooth. 
\begin{equation}
   \sum\limits_i {\min {{\left\| {{{\mathbf{S}}_i}} \right\|}_0}} \quad s.\:t.{\text{  }}\left\{ {\begin{array}{*{20}{c}}
  {{{\left\| {{\mathbf{AS}} - {\mathbf{Y}}} \right\|}_2} < \varepsilon } \\ 
  {priori{\text{ }}information} 
\end{array}} \right.
\label{eq4}
\end{equation}
\begin{equation}
    TV\left( {\mathbf{x}} \right) = \frac{1}{n}\sum\limits_{j = 1}^{n - 1} {{{\left( {{x_j} - {x_{j + 1}}} \right)}^2}} 
    \label{eq5}
\end{equation}

\begin{equation}
    TV\left( I \right) = \frac{1}{H}\sum\limits_{i = 1}^{H - 1} {{{\left( {{I_{i,j}} - {I_{i + 1,j}}} \right)}^2}}  + \frac{1}{W}\sum\limits_{j = 1}^{W - 1} {{{\left( {{I_{i,j}} - {I_{i,j + 1}}} \right)}^2}} 
    \label{eq6}
\end{equation}

The local variation is for images and high-dimensional signals, the motivation is a local region is smooth, such as a small block in images. LV is defined in Eq.(\ref{eq7}) as,
\begin{equation}
    LV\left( I \right) = \frac{1}{B}\sum\limits_{b = 1}^B {std\left( {{I_b}} \right)}
    \label{eq7}
\end{equation}
where $B$ is the segmented block numbers for the high-dimensional signal, $std(\cdot)$ is the standard deviation. For ease of implementation, a fixed-size sliding window can be used for block segmentation. The sliding step-size should smaller than the window size to prevent block effects in the reconstruction results. The default window size is 3$\times$3 to balance between image details and local smooth.

\begin{table}[tb]
  \centering
  \caption{\bf $\bf{s}$ forward block for some sparse reconstruction methods}
  \resizebox{1\textwidth}{!}{
    \begin{tabular}{lll}
    \toprule
    \textbf{Algorithms} & \boldmath{}\textbf{$\bf{s}$ forward}\unboldmath{} & \textbf{Note} \\
    \midrule
    OMP, SP, CoSaMP & ${\mathbf{s}} = pinv\left( {{\mathbf{A}}\left( {:,Mask} \right)} \right)$ & $Mask$ is a binary sequence derived from ${{\mathbf{A}}^T}{\mathbf{r}}$ to indicate the sub-columns \\
    \midrule
    IHT & ${\mathbf{s}} = {H_s}\left( {{\mathbf{s}} + t \cdot {{\mathbf{A}}^T}{\mathbf{r}}} \right)$ & $H_s$ is a thredholding function to select the largest values, $t$ is a scale factor  \\
    \midrule
    ISTA & ${\mathbf{s}} = sof{t_\lambda }\left( {{\mathbf{s}} - 2t \cdot {{\mathbf{A}}^T}{\mathbf{r}}} \right)$ & $sof{t_\lambda }$ is a soft threshold function, $t$ is a scale factor \\
    \midrule
    IRLS & ${\mathbf{s}} = inv{\left( {{{\mathbf{A}}^T}{\mathbf{WA}} + \lambda {\mathbf{I}}} \right)}{{\mathbf{A}}^T}{\mathbf{Wy}}$ & $\bf{W}$ is an adjustable weighting matrix, $\lambda$ is a scale factor \\
    \bottomrule
    \end{tabular}%
    }
  \label{tabAlg}%
\end{table}%

\subsection*{The Ultra Small Model --- Coefficients Learning}
\subsubsection*{General idea of coefficients learning}
\begin{figure}[!t]
\centering
\includegraphics[width=\linewidth]{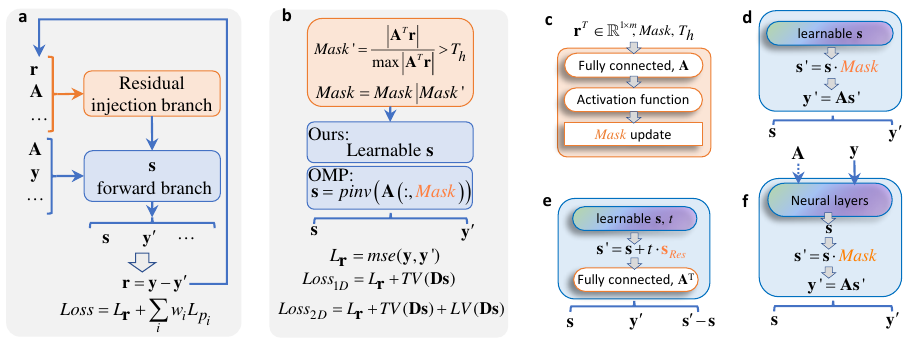}
\caption{\textbf{Diagram of the proposed ultra small artificial neural models---coefficients learning (CL) for training-free and universal sparse reconstruction.} {($\bf{a}$) CL reforms traditional iterative methods into one residual injection branch and one $\bf{s}$ forward branch with learnable $\bf{s}$, so $\bf{s}$ is learned with the residual and prior loss, thereby eliminating the need for a large-scale training dataset while maintaining generality and interpretability. ($\bf{b}$) shows the block function for CLOMP and general prior loss for 1D and 2D signals, where the learnable $s$ is used to substitute the psduo-inverse operation. More implementation structures for the two branches are provided in ($\bf{c-f}$). Since $\mathbf{s} \in {\mathbb{R}^{n \times 1}}$, the structures in ($\bf{c-e}$) only need a minimal of $n$ trainable parameters. These structures support parallel processing, so CL works for both one-dimensional and high-dimensional data.}}
\label{figMethod}
\end{figure}

Many iterative reconstruction algorithms can be divided into two blocks, the first block injects some new information from the residual, while the second block forward current $\bf{s}$ and new residual information to update the estimation. The residual injection block typically utilizes the correlation between the residual and each column of $\bf{A}$ to get new information from the residual, i.e., ${{\mathbf{A}}^T}{\mathbf{r}}$, such as OMP, IHT, ISTA, subspace pursuit (SP), CoSaMP, etc. Instead of calculating the residual by $\bf{r}={\mathbf{As}} - {\mathbf{y}}$, some methods uses $\bf{y}$ and some adjustable parameters. For example, iteratively reweighted least squares (IRLS) computes the residual as $\bf{Wy}$, where $\bf{W}$ is a weighting matrix that is adjusted in each iteration. No matter how the residual is calculated, since $\bf{A}$ is fixed, ${{\mathbf{A}}^T}{\mathbf{r}}$ can be implemented with artificial neural network structure, as shown in Figure \ref{figMethod}c, where $\bf{A}$ or $\bf{A}^T$ becomes a fixed parameter of a fully connection layer (or other structure like convolutional layer). Additional operations for ${{\mathbf{A}}^T}{\mathbf{r}}$ can be expressed as activation functions following the fully connected layer. Therefore, the first core idea of coefficients learning is reforming the residual injection into an artificial neural network structure with fixed parameters for $\bf{A}$ or $\bf{A}^T$ to expedite the calculation. which is called residual injection branch, as illustrated in Figure \ref{figMethod}a.

The $\bf{s}$ forward block relies on different methods, some popular methods are listed in Table \ref{tabAlg}. These methods typically involve operations with high computational complexity, such as matrix inversion, pseudo-inversion, and threshold functions, which can be quite intricate. Therefore, the second core idea of coefficients learning is to make $\bf{s}$ becomes learnable parameters, rather than relying on the calculation methods in Table \ref{tabAlg}, which is called $\bf{s}$ forward branch. $\bf{s}$ are automatically learned with the help of residual injection branch and the loss function, i.e., the residual loss $L_{\mathbf{r}}$ and the prior information loss $\sum_i {{w_i}{L_{{p_i}}}} $, where $w_i$ are scale factors for different prior terms.

The general framework of coefficients learning is summarized in Figure \ref{figMethod}a, where the residual injection branch takes inputs such as $\bf{r}$, $\bf{A}$, or other parameters like $\bf{Wy}$, and produces information to help $\bf{s}$ forward. The $\bf{s}$ forward branch can also takes these parameters as input or operate without any input. The outputs are at least $\bf{s}$ and the estimation of ${\mathbf{y'}} = {\mathbf{As}}$. Some implementation structures for $\bf{s}$ forward branch are illustrated in Figure \ref{figMethod}(d-f). In Figure \ref{figMethod}(d,e), $\bf{s}$ is directly set as learnable parameters, another option is Figure \ref{figMethod}f, where a neural block can be employed that takes fixed input such as $\bf{y}$ or $\bf{y}$ with $\bf{A}$, and utilizes various neural layers to output $\bf{s}$. To implement the 0-norm in Eq.(\ref{eq4}), the residual injection branch outputs a $Mask$ that indicates which positions are involved in the calculations for support-based methods such as OMP, SP, CoSaMP, and others. This mask is multiplied with $\bf{s}$ before being linked to the subsequent layers. For addition-based methods like IHT and ISTA, as illustrated in Figure \ref{figMethod}e, the learnable $\bf{s}$ can be updated by minimizing ${\mathbf{s'}} - {\mathbf{s}}$. Algorithms like IRLS do not require the sparse positions of $\bf{s}$, in this case, both $\bf{s}$ and $\bf{W}$ can be set as learnable parameters to supervise each other mutually. During coefficients learning, any prior information can be incorporated into the loss function, resulting in improved accuracy compared to the parent algorithm.

\subsubsection*{Coefficients learning on OMP (CLOMP)}
Taking OMP as an example, this section shows how OMP algorithm can be reformulated within the proposed CL framework, thereby bypassing the complex pseudo-inverse operation. OMP identifies additional columns in $\bf{A}$ that exhibit a high correlation with the residual. Subsequently, the coefficients are computed using the pseudo-inverse of all the highly correlated sub-columns of $\bf{A}$. The positions of these sub-columns correspond to the locations of the non-zero coefficients in $\bf{s}$. So, it is efficient when the signal is sufficiently sparse, as the pseudo-inverse is applied only to the sub-columns.

\begin{figure}[tb]
\centering
\includegraphics[width=0.8\linewidth]{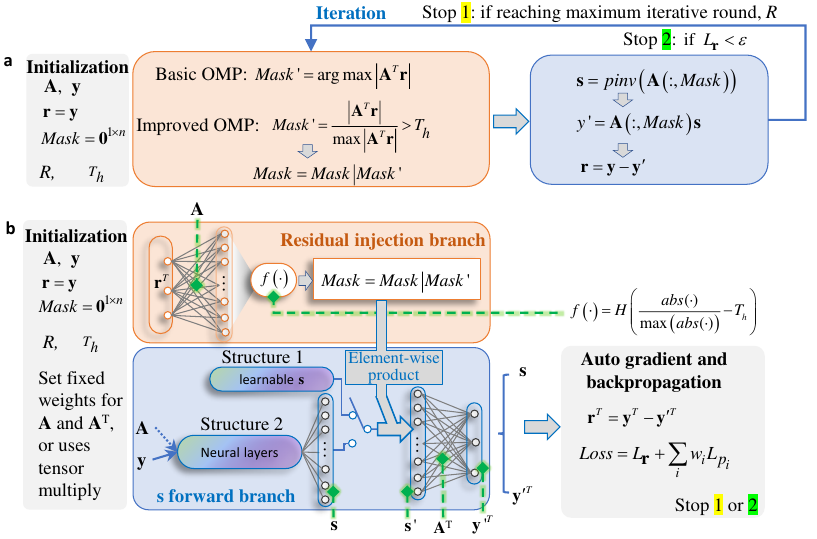}
\caption{\textbf{Structure of the proposed coefficients learning on OMP (CLOMP).} {The process for OMP is shown in ($\bf{a}$). ($\bf{b}$) shows the structure for the residual injection branch and $\bf{s}$ forward branch for CLOMP, where $H(x)$ is the unit step function which is 0 for $x$>0, otherwise 1.}}
\label{figCLOMP}
\end{figure}

As shown in Figure \ref{figCLOMP}a for OMP, the residual injection block calculate ${{\mathbf{A}}^T}{\mathbf{r}}$ first in each iteration. This process can be reformulated into an artificial neural network structure, as illustrated in the residual injection branch in Figure \ref{figCLOMP}b, where $\bf{r}$ is transposed to facilitate parallel computing, $\bf{A}$ is set as fixed parameters for a fully connected layer during initialization. The basic OMP only identifies the maximum positions in ${{\mathbf{A}}^T}{\mathbf{r}}$, which is relatively slow. The improved OMP, sometimes referred to as stOMP in the literature, identifies multiple positions in a single iteration instead. After passing ${\bf{r}}^T$ through this layer, ${{\mathbf{A}}^T}{\mathbf{r}}$ is computed. Next, the absolute operation and normalization process can be expressed as an activation function $f(\cdot)$, or using $softmax$ directly after taking the absolute value. The mask update is performed using a bitwise $or$ operation between the current and previous mask, which is also very fast. For the $\bf{s}$ forward branch, the two structures depicted in Figure \ref{figCLOMP}b can be used. We find that structure 1 is more efficient and accurate than structure 2; therefore, the results section employs structure 1. Structure 2 has more learnable parameters, which may enhance accuracy in applications with more prior information. Next, $\bf{s}$ is multiplied by the mask in an element-wise manner to implement the 0-norm. Subsequently, another fully connected layer that uses $\bf{A}^T$ as fixed parameters can be employed to replace the $\bf{y}' = {\mathbf{As}}'$. Finally, the loss function in \ref{figMethod}b can be applied to 1D and 2D signals. With the auto-gradient and backpropagation capabilities of neural networks, $\bf{s}$ is learned automatically.

With the design outlined above, CLOMP accepts the same input as OMP and outputs $\bf{s}$ or $\bf{x}$, $\bf{s}$ is learned during the iteration without the need for model training. CLOMP is applicable to any type of signal and any scenario, similar to OMP, which ensures its perfct generality. Since the structure of CLOMP is based on OMP, there is no black-box, so the iterpretability is also perfect. CLOMP can incorporate prior information into the loss function, a capability that surpasses traditional iterative algorithms, resulting in significantly improved accuracy. Its parallel processing nature and the absence of a pseudo-inverse design contribute to CLOMP's speed, particularly when handling large-scale data.

\bibliography{sample.bib}

\section*{Acknowledgements}
This work is partially funded by National Natural Science Foundation of China
(no.62103154); Young Elite Scientists Sponsorship Program by CAST (no.2023QNRC001); National Science Fund for Distinguished Young Scholars of China (no.62225603); Key International Cooperation Research Projects of NSFC (no.61960206010).

\section*{Author contributions statement}
Conceived and designed the experiments, C.T., G.T, Y.D., W.L., X.B.; Performed the experiments, C.T., H.Z.,Z.Z.; Analyzed the data, C.T., H.Z, Z.Z., L.L.; Contributed materials/analysis tools, C.T., H.Z., Y.D., W.L., X.B.; Wrote the paper, C.T., G.T., X.B.

\section*{Data and code availability}
 The data and code for this paper are available at \url{https://github.com/BillTtzqgbt/CSCoefficientsLearning.git}.

\section*{Competing interests}
The authors declare no competing interests.

\section*{Additional information}

\textbf{Appendix figures}:
\appendix

\setcounter{figure}{0}
\renewcommand{\thefigure}{A\arabic{figure}}

\begin{figure}[htb]
\centering
\includegraphics[width=\linewidth]{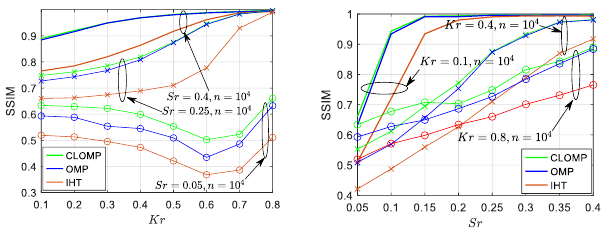}
\caption{\textbf{Reconstruction accuracy on SSIM versus $Kr$ (the left) and $Sr$ (the right) on synthetic 1D signals.}}
\label{figA1}
\end{figure}

\begin{figure}[htb]
\centering
\includegraphics[width=\linewidth]{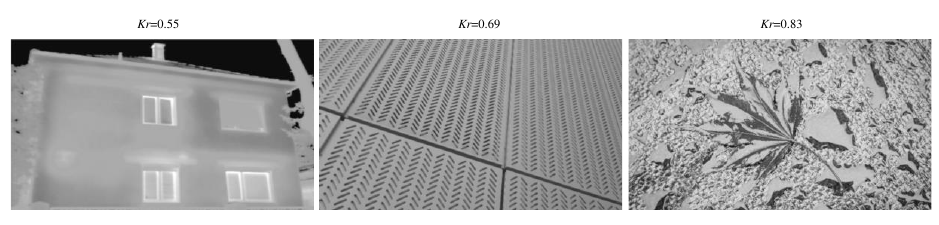}
\caption{\textbf{A group of source images for synthetic 2D signals.}}
\label{figA2}
\end{figure}

\end{document}